\documentclass[journal]{IEEEtran}

\usepackage{bbm}
\usepackage{caption}
\usepackage{amsthm}
\usepackage{amssymb,graphicx,lineno}
\usepackage{rawfonts,amsfonts,amssymb,psfrag,color}
\usepackage{amsfonts}
\usepackage{mathrsfs}
\usepackage{amsmath,epsfig}
\usepackage{multirow}
\usepackage{stfloats}
\usepackage{rawfonts,graphicx,amsfonts,amssymb}
\usepackage{bm}
\usepackage{algorithm, algorithmic}
\usepackage{verbatim}  

\usepackage{subfigure}
\usepackage{enumerate}
\usepackage{cooltooltips}
\usepackage[dvipsnames]{xcolor}

\usepackage{datetime}
\newdateformat{ukdate}{}

\begin{document}

\title{Nonlinear Tensor Ring Network}

\author{Xiao Peng~Li,~Qi~Liu, \IEEEmembership{Member, IEEE},~Hing Cheung~So,~\IEEEmembership{Fellow,~IEEE}

        \thanks{Xiao Peng Li and Hing Cheung So are with the Department of Electrical
        Engineering, City University of Hong Kong, Hong Kong SAR, China (e-mail:
        x.p.li@my.cityu.edu.hk; hcso@ee.cityu.edu.hk). (Corresponding author: Qi~Liu.)}
        \thanks{Q. Liu is with the Department of Electrical and Computer Engineering, National University
        of Singapore, Singapore (e-mail: elelqi@nus.edu.sg).}
}

\markboth{\ukdate\today}%
{Shell \MakeLowercase{\textit{et al.}}: Bare Demo of IEEEtran.cls for IEEE Journals}

\maketitle

\begin{abstract}
The state-of-the-art deep neural networks (DNNs) have been widely applied for various real-world applications, and achieved significant performance for cognitive problems. However, the increment of DNNs' width and depth in architecture results in a huge amount of parameters to challenge the storage and memory cost, limiting to the usage of DNNs on resource-constrained platforms, such as portable devices. By converting redundant models into compact ones, compression technique appears to be a practical solution to reducing the storage and memory consumption.  
In this paper, we develop a nonlinear tensor ring network (NTRN) in which both fully-connected and convolutional layers are compressed via tensor ring decomposition. Furthermore, to mitigate the accuracy loss caused by compression, a nonlinear activation function is embedded into the tensor contraction and convolution operations inside the compressed layer. Experimental results demonstrate the effectiveness and superiority of the proposed NTRN for image classification using two basic neural networks, LeNet-5 and VGG-11 on three datasets, viz. MNIST, Fashion MNIST and Cifar-10.

\end{abstract}


\IEEEpeerreviewmaketitle

\section{Introduction}\label{sec: introduction}
\IEEEPARstart{R}{ecently},  deep neural networks (DNNs) have attracted increasing attention due to their excellent performance in various fields, such as speech recognition~\cite{xu2013experimental,liu2020parameter}, image denoising~\cite{wang2021channel}, video categorization~\cite{jiang2017exploiting}, object detection~\cite{guo2020doa} and bioinformatics~\cite{li2019deep}.
A pivotal characteristic of the DNNs is large depth since increasing the number of layers is able to improve the representational ability for attaining higher accuracy~\cite{he2016deep}. The vast depth generates an enormous amount of parameters that requires huge storage and memory space, restricting the DNN deployment on resource-constrained devices, such as mobile phones, wearables and internet of things (IoTs). To that end, network compression technique~\cite{han2016deep} is proposed to compress DNNs. 

In general, the methods for neural network compression can be roughly classified into four categories, namely, parameter pruning and quantization, low-rank factorization, transferred/compact convolution filters, and knowledge distillation~\cite{cheng2017survey}. Compared with the others, low-rank factorization is easy to implement for supporting scratch and pre-trained neural networks because of its standardized pipeline. Thereby, in this work, we focus on low-rank factorization that leverages the low-rank property in matrix/tensor to decompose a weight array into multiple small-size factors for reducing the number of parameters. Low-rank matrix decomposition was proposed to compress multi-layer perceptrons (MLPs)~\cite{sainath2013low} and convolutional neural networks (CNNs)~\cite{tai2015convolutional}. However, the compressed models inevitably suffer from the performance degradation. To alleviate the loss of accuracy at high compression ratio (CR),~\cite{yu2017compressing} suggested to combine the low-rank and sparse decomposition to compress DNNs. The joint strategy factorizes a weight matrix into two small-size factor matrices and one sparse matrix based on the pre-trained neural network, and then fine-tunes the matrices to improve performance. In addition, neural network compression was considered as a multi-objective optimization problem in~\cite{huang2020deep}, where CR and classification error ratio were optimized simultaneously. Although it can achieve satisfactory tradeoff between CR and classification error ratio, compression artifacts may exist in low-rank matrix decomposition based approaches because they have to reshape the tensorial weights in convolutional layers into a matrix.

To avoid impairing tensor structure, low-rank tensor decomposition is utilized to compress DNNs, and has achieved higher CR and accuracy compared with the counterparts. The prior work to compress DNNs was using Tucker decomposition~\cite{kim2015compression}, followed by the CANDECOMP/PARAFAC (CP) decomposition~\cite{astrid2017cp} to reduce performance loss at high CR. Besides, tensor train (TT)~\cite{garipov2016ultimate} and tensor ring (TR)~\cite{wang2018wide,zhao2019learning} models were utilized for neural network compression. They convert the weight array into a high order tensor for a high CR, and then factorize the corresponding tensor into multiple core tensors. 
Specifically, TR achieved a 18-times (18x) CR with only 0.1\% accuracy loss on LeNet-5 model for processing MNIST dataset. Furthermore, Sun $et$ $al.$~\cite{sun2020deep} leveraged the structure sharing feature in ISTA-Net~\cite{zhang2018ista} and ResNet~\cite{he2016deep} to achieve a high CR with small accuracy loss. Recently, a nonlinear TT (NTT) method~\cite{wang2021nonlinear} was developed to insert a dynamic activation function between two adjacent core tensors to boost performance.


In this work, we aim to leverage low-rank tensor factorization to compress DNNs. It has been demonstrated that TR format is able to achieve higher accuracy than TT decomposition at the same CR~\cite{pan2019compressing}. Thereby, we adapt TR factorization to devise a nonlinear TR network (NTRN), where both fully-connected and convolutional layers are compressed by TR factorization. Besides, a nonlinear activation function is added after tensor contraction and convolution operations inside the compressed layer. Different from the NTT format that unfolds the tensorial weights in convolutional layers into a matrix, the proposed NTRN takes action directly on tensorial weights, without distorting the spatial information in convolution kernels. To verify the expressive ability of the proposed NTRN, we test the NTRN with two basic neural networks, LeNet-5~\cite{lecun1998gradient} and VGG-11~\cite{Simonyan2015} on MNIST~\cite{lecun1998gradient}, Fashion MNIST~\cite{xiao2017fashion} and Cifar-10~\cite{Krizhevsky2009} datasets. The experimental results demonstrate that our NTRN is able to alleviate the accuracy loss compared with the TR based method.

\section{Preliminaries}\label{sec: preliminaries}
Consider a $N$th-order tensor $\pmb{\mathcal{A}}\in \mathbb{R}^{I_1\times \cdots\times I_N}$, TT decomposition represents $\pmb{\mathcal{A}}$ by two matrices and $N-2$ core tensors, that is
\begin{align}
	\pmb{\mathcal{A}}(i_1,i_2,\!\cdots\!,&i_n)\! = \!\!\sum_{r_1,r_2,\!\cdots\!,r_{N-1}=1}^{R_1,R_2,\!\cdots\!,R_{N-1}}\! \pmb{G}_1(i_1,\!r_1)\pmb{\mathcal{G}}_2(r_1,\!i_2,\!r_2)\!\cdots\! \nonumber\\
	&\pmb{\mathcal{G}}_{N-1}(r_{N-2},\!i_{N-1},\!r_{N-1})\pmb{G}_{N}(r_{N-1},\!i_{N})
\end{align}
where $\pmb{\mathcal{A}}(i_1,i_2,\!\cdots\!,i_n)$ denotes the $(i_1,i_2,\!\cdots\!,i_n)$ entry, $\pmb{G}_1\in \mathbb{R}^{I_1\times R_1}$, $\pmb{G}_N\in \mathbb{R}^{R_{N-1}\times I_N}$, $\pmb{\mathcal{G}}_k\in \mathbb{R}^{R_{k-1}\times I_k\times, R_k}$ with $k\in [2,N-1]$ are termed as core tensors, and $[R_1,R_2,\cdots,R_{N-1}]$ are TT ranks. By contrast, TR decomposes $\pmb{\mathcal{A}}$ into $N$ 3rd-order tensors, shown as
\begin{align}
	\pmb{\mathcal{A}}(i_1,i_2,\!\cdots\!,&i_n)\! =\! \!\sum_{r_1,r_2,\!\cdots\!, r_N=1}^{R_1,R_2,\!\cdots\!,R_{N}}\! \pmb{\mathcal{G}}_1(r_1,i_1,\!r_2)\pmb{\mathcal{G}}_2(r_2,\!i_2,\!r_3)\!\cdots\! \nonumber\\
	&\pmb{\mathcal{G}}_{N-1}(r_{N-1},\!i_{N-1},\!r_{N})\pmb{\mathcal{G}}_{N}(r_{N},\!i_{N},r_{1})
\end{align}
where $\pmb{\mathcal{G}}_k\in \mathbb{R}^{R_k\times I_k\times R_{k+1}}$ with $k\in [1,N]$ and $[R_1,R_2,\cdots,R_{N-1},R_N]$ are TR ranks. The main difference between TT and TR factorization is in the first and last core arrays. Specifically, the TT format requires two matrices in the first and last positions, while all core arrays are 3rd-order tensors in the TR factorization. Thereby, the TT decomposition might result in large intermediate cores and small boundary factors, which restricts its representational ability and flexibility. Besides, multiplying TT cores must obey a strict order, and hence the convolution kernel in convolutional layers is unfolded into a vector for performing sequenced operation~\cite{garipov2016ultimate}. This motivates us to devise a nonlinear TR format for neural network compression.

\section{Nonlinear Tensor Ring Network}\label{sec: proposedModel}
In this section, the proposed NTRN to compress fully-connected and convolutional layers is introduced, respectively, in detail.
\subsection{Fully-Connected Layer Decomposition}
\begin{figure}[!htb]
	\centering
	\includegraphics[width=3in]{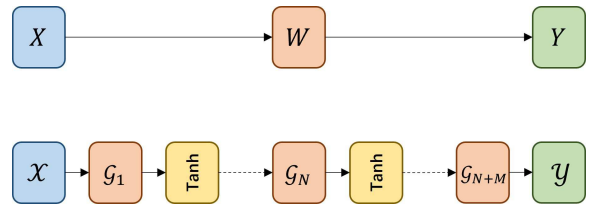}
	\caption{Illustrations of fully-connected layer in conventional neural network and nonlinear tensor ring network. }
	\label{fig: FCL}
\end{figure}

A fully-connected layer maps an input vector $\pmb x\in \mathbb{R}^{I}$ into an output vector $\pmb y \in \mathbb{R}^{O}$ via a weight matrix $\pmb W\in \mathbb{R}^{I\times O}$. Mathematically, it is formulated as
\begin{equation}\label{eq: FCL}
	\begin{aligned}
		\pmb y = \pmb W^T\pmb x
	\end{aligned}
\end{equation}
where $(\cdot)^T$ denotes the transpose operator. Neural network compression by low-rank matrix/tensor factorization is to decompose $\pmb W$ into some small-size factors. To compress $\pmb W$ in tensor structure, we first reshape $\pmb W$ into a high order tensor $\pmb{\mathcal{W}}\in \mathbb{R}^{I_1\times \cdots\times I_N\times O_1\times \cdots\times O_M}$ with 
\begin{equation}
	\begin{aligned}
		\prod_{n=1}^{N}I_n = I, \prod_{m=1}^{M}O_m = O.
	\end{aligned}
\end{equation}
Then, based on TR format, $\pmb{\mathcal{W}}$ is factorized into $(N+M)$ 3rd-order tensors
\begin{align}\label{TRD}
	\pmb{\mathcal{W}}= \sum\limits_{r_1,\cdots,r_n = 1}^{R_1,\cdots,R_{M+N}}&\pmb{\mathcal{G}}_1(r_1,:,r_{2})\circ \pmb{\mathcal{G}}_2(r_2,:,r_{3})\circ\cdots\nonumber\\
	&\circ \pmb{\mathcal{G}}_{M+N}(r_{M+N},:,r_{1})
\end{align}
where $\circ$ implies the outer product operation, $\pmb{\mathcal{G}}_k\in \mathbb{R}^{R_k\times I_k\times R_{k+1}}$ with $k\in[1,N]$, $\pmb{\mathcal{G}}_k\in \mathbb{R}^{R_k\times O_k\times R_{k+1}}$ with $k\in[N+1,M]$, and $\pmb{\mathcal{G}}_k(r_k,:,r_{k+1})$ is the vertical fiber of $\pmb{\mathcal{G}}_k$.
Moreover, $\pmb x$ and $\pmb y$ need to be tensorized as $\pmb{\mathcal{X}} \in \mathbb{R}^{I_1\times \cdots\times I_N}$ and $\pmb{\mathcal{Y}} \in \mathbb{R}^{O_1\times \cdots\times O_M}$, respectively. Thereby,~\eqref{eq: FCL} can be rewritten in the tensor format:
\begin{align}\label{eq: TR_FCL}
	\pmb{\mathcal{Y}} = &\pmb{\mathcal{X}}\times_{1}^{2}\pmb{\mathcal{G}}_1 \times_{1,N+1}^{2,1}\pmb{\mathcal{G}}_2 \cdots \times_{1,3}^{2,1}\pmb{\mathcal{G}}_N \times_{2}^{1}\pmb{\mathcal{G}}_{N+1} \cdots \nonumber\\
	&\times_{M-1}^{1}\pmb{\mathcal{G}}_{N+M-1} \times_{1, M+1}^{3,1}\pmb{\mathcal{G}}_{N+M}
\end{align} 
where $\times_{k}^{1}$ and $\times_{k,l}^{1,3}$ are the tensor contraction operations~\cite{cichocki2016low}. Consider $\pmb{\mathcal{A}}\in \mathbb{R}^{I_1\times \cdots\times I_{k-1} \times I_k \times I_{k+1} \times \cdots \times I_{l-1} \times I_l \times I_{l+1}\times \cdots \times I_N}$ and $\pmb{\mathcal{B}}\in \mathbb{R}^{J_1\times J_{2} \times J_3}$ with $I_k = J_1$ and $I_l=J_3$, $\pmb{\mathcal{A}} \times_{k}^{1} \pmb{\mathcal{B}}$ yields a $(N+1)$th-order tensor $\pmb{\mathcal{C}}\in \mathbb{R}^{I_1\times \cdots\times I_{k-1} \times I_{k+1} \times \cdots \times I_N\times J_2 \times J_3}$ whose entries are calculated by
\begin{align}
	\pmb{\mathcal{C}}(i_1, \cdots, i_{k-1}, &i_{k+1}, \cdots i_N, j_2 , j_3) = \nonumber\\
	&\sum_{i_k= 1}^{I_k}\pmb{\mathcal{A}}(i_1, \cdots,i_k, \cdots i_N)\pmb{\mathcal{B}}(i_k,j_2,j_3).
\end{align} 
Similarly, $\pmb{\mathcal{A}} \times_{k,l}^{1,3} \pmb{\mathcal{B}}$ generates a $(N-1)$th-order tensor $\pmb{\mathcal{D}}\in \mathbb{R}^{I_1\times \cdots\times I_{k-1} \times I_{k+1} \times \cdots \times I_{l-1} \times I_{l+1}\times \cdots\times I_{N}  \times J_{2}}$ with entries being
\begin{align}
	\pmb{\mathcal{D}}(i_1, \cdots, i_{k-1}, i_{k+1}, \cdots, i_{l-1}, i_{l+1}, \cdots, i_{N}, j_{2}) = \nonumber \\
	\sum_{i_k, i_{l}= 1}^{I_k, I_l}\pmb{\mathcal{A}}(i_1, \cdots, ,i_k, \cdots, i_{l}, \cdots i_N)\pmb{\mathcal{B}}(i_k,j_2,i_{l}).
\end{align} 
It is worth noting that~\eqref{eq: TR_FCL} is the same as the fully-connected TR network in~\cite{wang2018wide}. To boost the accuracy, we propose to include a nonlinear activation function after each tensor contraction in~\eqref{eq: TR_FCL}, resulting in
\begin{align}\label{eq: NTR_FCL}
	\pmb{\mathcal{Y}} = &f(\cdots f(f( \cdots f(\pmb{\mathcal{X}}\times_{1}^{2}\pmb{\mathcal{G}}_1)  \cdots \times_{1,3}^{2,1}\pmb{\mathcal{G}}_N)  \nonumber\\
	& \times_{2}^{1}\pmb{\mathcal{G}}_{N+1}) \cdots \times_{M-1}^{1}\pmb{\mathcal{G}}_{N+M-1}) \times_{1, M+1}^{3,1}\pmb{\mathcal{G}}_{N+M}
\end{align} 
where $f(\cdot)$ signifies the nonlinear activation function, e.g., Tanh in Fig.~\ref{fig: FCL}. Note that $f(\cdot)$ is not added after $\pmb{\mathcal{G}}_{N+M}$ since the intrinsic activation function outside the current layer replaces $f(\cdot)$. Fig.~\ref{fig: FCL} illustrates the fully-connected layer in NTRN.

\subsection{Convolutional Layer Decomposition}
\begin{figure}[!htb]
	\centering
	\includegraphics[width=3in]{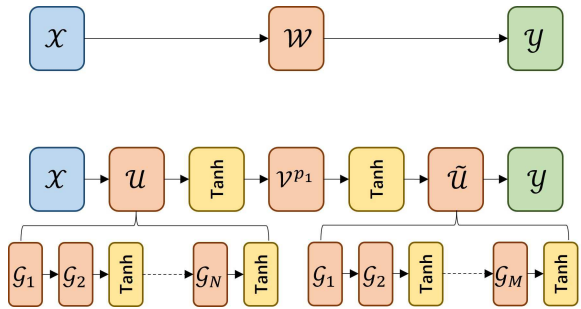}
	\caption{Illustrations of convolutional layer in conventional neural network and nonlinear tensor ring network. }
	\label{fig: CL}
\end{figure}

At the convolutional layer, an input tensor $\pmb{\mathcal{X}}\in \mathbb{R}^{H\times W\times I}$ is convoluted with a 4th-order kernel tensor $\pmb{\mathcal{W}}\in \mathbb{R}^{D\times D\times I\times O}$ to output a tensor $\pmb{\mathcal{Y}}\in \mathbb{R}^{\widetilde{H}\times \widetilde{W}\times O}$, formulated as
\begin{equation}\label{eq: convolutionalOperation}
	\pmb{\mathcal{Y}} = \pmb{\mathcal{W}} * \pmb{\mathcal{X}}
\end{equation}
where $*$ denotes the convolution operation in CNNs. To compress $\pmb{\mathcal{W}}$, TR decomposition factorizes it into three core tensors, viz. $\pmb{\mathcal{V}}\in \mathbb{R}^{R_1\times D\times D\times R_2}$, $\pmb{\mathcal{U}}\in \mathbb{R}^{R_2\times I\times R_3}$ and $\pmb{\mathcal{\widetilde{U}}}\in \mathbb{R}^{R_3\times O\times R_1}$. Therefore, $\pmb{\mathcal{W}}$ is computed by
\begin{equation}
	\begin{aligned}
		 \pmb{\mathcal{W}} 
		 = \sum_{r_1,r_2,r_3 = 1}^{R_1,R_2,R_3}\pmb{\mathcal{V}}(r_1,:,:,r_2)\circ \pmb{\mathcal{U}}(r_2,:,r_3)\circ \pmb{\mathcal{\widetilde{U}}}(r_3,:,r_1)
	\end{aligned}
\end{equation}
It can be known that $\pmb{\mathcal{V}}$ retains convolution kernel, which is different from NTT format.
In NTRN, the convolution operation~\eqref{eq: convolutionalOperation} is decomposed into the following procedure:
\begin{align}
	\pmb{\mathcal{Z}}_1 &= f(\pmb{\mathcal{X}} \times_{3}^{2} \pmb{\mathcal{U}})\\
	\pmb{\mathcal{Z}}_2 &= f(\pmb{\mathcal{V}}^{p_1} * \pmb{\mathcal{Z}}_1)\\
	\pmb{\mathcal{Y}} &= \pmb{\mathcal{Z}}_2 \times_{1, 3}^{3, 1} \pmb{\mathcal{\widetilde{U}}}
\end{align}
where $\pmb{\mathcal{Z}}_1\in \mathbb{R}^{H\times W\times R_2\times R_3}$, $\pmb{\mathcal{Z}}_2\in \mathbb{R}^{\widetilde{H}\times \widetilde{W} \times R_1\times R_3}$ and $(\cdot)^{p_1}$ is the 1st tensor permutation. The relationship between $\pmb{\mathcal{V}}^{p_1}$ and $\pmb{\mathcal{V}}$ obeys
\begin{equation}
	\pmb{\mathcal{V}}^{p_1}(d_1,d_2,\alpha_2,\alpha_1) = \pmb{\mathcal{V}}(\alpha_1, d_1,d_2,\alpha_2).
\end{equation}
For the convolutional layer in NTRN, the input tensor $\pmb{\mathcal{X}}$ is first contracted with the intermediate factor $\pmb{\mathcal{U}}$, then convoluted with the first core $\pmb{\mathcal{V}}$, and finally contracted with the last core $\pmb{\mathcal{\widetilde{U}}}$. It is worth mentioning that this procedure is entirely different from the operation in NTT format.
On the other hand, when $I$ and $O$ are large, $\pmb{\mathcal{U}}$ and $\pmb{\mathcal{\widetilde{U}}}$ can be further factorized into small-size core tensors for achieving a higher CR. For instance, $\pmb{\mathcal{U}}$ can be reshaped as $\pmb{\mathcal{\widehat{U}}} \in \mathbb{R}^{R_2\times I_1\times I_2\times I_3\times R_3}$ with $I_1\times I_2\times I_3 = I$, and then $\pmb{\mathcal{\widehat{U}}} $ is factorized into three tensors, namely, $\pmb{\mathcal{G}}_1 \in \mathbb{R}^{R_2\times I_1\times R_{22}}$, $\pmb{\mathcal{G}}_2 \in \mathbb{R}^{R_{22}\times I_2\times R_{23}}$ and $\pmb{\mathcal{G}}_3 \in \mathbb{R}^{R_{23} \times I_3\times R_3}$. To calculate $\pmb{\mathcal{\widehat{U}}}$ from its core tensors, the proposed NTRN adopts a nonlinear contraction operation
\begin{equation}
	\pmb{\mathcal{\widehat{U}}} = f(f(\pmb{\mathcal{G}}_1 \times_{3}^{1} \pmb{\mathcal{G}}_2)\times_{4}^{1} \pmb{\mathcal{G}}_3).
\end{equation}
The convolutional layer in NTRN is illustrated in Fig.~\ref{fig: CL}.

\subsection{Compression Ratio}
CR is defined as:
\begin{equation}
	\text{CR} = \frac{\text{\# parameters of original network}}{\text{\# parameters of compressed network}}.
\end{equation}
For a fully-connected layer, the number of parameters is $I O$ given a weight matrix $\pmb W\in \mathbb{R}^{I\times O}$. The proposed NTRN requires $(\sum_{n = 1}^{N}I_n + \sum_{m = 1}^{M}O_m)R^2$ parameters with the assumption of $R_1=\cdots=R_{M+N}=R$. On the other hand, the convolutional layer of conventional neural network needs to store $D^2 I O$ parameters for $\pmb{\mathcal{W}}\in \mathbb{R}^{D\times D\times I\times O}$. In NTRN, the dimensions of $\pmb{\mathcal{W}}$ are reshaped into ${D\times D\times I_1\times \cdots\times I_N \times O_1\times \cdots\times O_M}$ for a high CR. Hence, only $(\sum_{n = 1}^{N}I_n + \sum_{m = 1}^{M}O_m + D^2)R^2$ parameters are required. It is known that decreasing $R$ is able to increase CR.

\section{Experimental Results}\label{sec: experiment}
In this section, the proposed NTRN is evaluated via basic MLP and CNN, LeNet-5~\cite{lecun1998gradient}, and VGG-11~\cite{Simonyan2015} on MNIST~\cite{lecun1998gradient}, Fashion MNIST~\cite{xiao2017fashion} and Cifar-10~\cite{Krizhevsky2009} datasets. All networks are programmed on PyTorch framework~\cite{paszke2019pytorch}. The experiments for the MLP, CNN and LeNet-5 are implemented on Nvidia GTX 2060 GPU, while VGG-11 is performed on Nvidia GTX 2080Ti GPU. 
In our experiments, TR ranks in an individual layer are the same, but can be diverse in different layers. If the activation function outside compressed layers, and the pooling function in CNNs are not specified, they are set as ReLU and maxpooling (size = 2, stride = 2), respectively.

\subsection{Fully-Connected Layer Evaluation}
\begin{table}[!htbp]
	\centering
	\caption{MLP compression on MNIST dataset.}
	\begin{tabular}{c c c c c c c c}\hline
		&  		& CR	& Acc  		& Std 		& H 	& L  	& Storage	\\ 	\hline\hline
		& Original 	& 1x	& 98.15  	& 0.001	 	& 98.3  & 98.0	& 5.1 MB	\\ 	\hline
		& TRN~\cite{wang2018wide}	& 57x  	& 97.34 	& 0.003		& 97.7	& 97.0	& 0.1 MB	\\	\hline
		& NTRN 	& 57x 	& 97.72 	& 0.002		& 97.9	& 97.4 	& 0.1 MB	\\	\hline
		& TRN~\cite{wang2018wide}  	& 359x  & 95.94  	& 0.004 	& 96.3 	& 95.2	& 0.02 MB	\\	\hline
		& NTRN  & 359x  & 96.51 	& 0.002		& 96.7  & 96.3	& 0.02 MB	\\	\hline
	\end{tabular}
	\label{table: MLP}
\end{table}

We first test the proposed NTRN using a MLP on MNIST dataset. The MLP consists of input, two hidden and output layers with 784, 1024, 512 and 10 nodes, respectively. All images in the MNIST are reshaped as a vector of length $784$. Therefore, the weight matrices of the MLP are $\pmb W_1\in \mathbb{R}^{784\times 1024}$, $\pmb W_2\in \mathbb{R}^{1024\times 512}$ and $\pmb W_3\in \mathbb{R}^{512\times 10}$. To compress them, $\pmb W_1$, $\pmb W_2$ and $\pmb W_3$ are tensorized, and hence their dimensions are $4\times 7\times 4\times 7 \times 4\times 8\times 4\times 8$, $4\times 8\times 4\times 8 \times 8\times 8\times 8$ and $8\times 8\times 8 \times 10$, respectively. Epoch is set as 50 for  training. The experimental results based on 20 independent trials are tabulated in Table~\ref{table: MLP}, where Acc and Std denote the average accuracy on the test set and the standard deviation of all results, respectively. In addition, the highest and lowest accuracy among all trials are listed in columns H and L, respectively. Moreover, storage indicates the disk space occupied by these parameters. At 57x CR, the ranks of three layers are $\{16, 14, 8\}$, while the ranks are set as $\{6, 5, 5\}$  359x CR. We can see that the proposed NTRN  outperforms TRN~\cite{wang2018wide} at both low and high CRs in terms of the average accuracy, the standard deviation, the highest and lowest accuracy. 

\subsection{Convolutional Layer Evaluation}
We now investigate the compression performance via a CNN with two hidden convolutional and one fully-connected layers on MNIST. The first and second convolutional layers contain 16 and 32 kernels with kernel-size equaling 3, respectively. Moreover, the stride is set to 1, and the padding is 0. Since the dimensions of input data are $28\times 28\times 1$, the weight tensor at the first convolutional layer is $\pmb{\mathcal{W}}_1\in \mathbb{R}^{3\times 3 \times 1\times 16}$. Besides, the weight arrays at the second convolutional and fully-connected layers are expressed as $\pmb{\mathcal{W}}_2\in \mathbb{R}^{3\times 3 \times 16\times 32}$ and $\pmb{\mathcal{W}}_3\in \mathbb{R}^{320\times 10}$, respectively. To compress weight arrays of the convolutional layers, the dimensions of $\pmb{\mathcal{W}}_1$ and $\pmb{\mathcal{W}}_2$ are reshaped as $3\times 3 \times 1 \times 4 \times 4$ and $3\times 3 \times 4 \times 4 \times 4 \times 8$, respectively. Following the ablation study, the fully-connected layer in CNN is not compressed. Hence, we do not resize its weight matrix.


The results are shown in Table~\ref{table: CNN}. The ranks at 4.3x and 8.9x CRs are $\{2, 6\}$ and $\{2, 4\}$, respectively. It is seen that the proposed NTRN attains higher accuracy than TRN at the same CR. While the standard deviation of NTRN is smaller than that of TRN. Note that the storage information is not provided as the fully-connected layer is not compressed. 

\begin{table}[!htbp]
	\centering
	\caption{CNN compression on MNIST dataset.}
	\begin{tabular}{c c c c c c c}\hline
		&  		& CR	& Acc  		& Std 		& H 	& L  		\\ 	\hline\hline
		& Original 	& 1x	& 98.71  	& 0.001	 	& 98.9  & 98.5		\\ 	\hline
		& TRN~\cite{wang2018wide}  	& 4.3x  & 98.35  	& 0.003 	& 98.6	& 98.1		\\	\hline
		& NTRN  & 4.3x  & 98.60 	& 0.001		& 98.9	& 98.4		\\	\hline
		& TRN~\cite{wang2018wide}  	& 8.9x  & 97.86  	& 0.003 	& 98.3	& 97.2		\\	\hline
		& NTRN  & 8.9x  & 98.15 	& 0.002		& 98.4	& 97.8		\\	\hline
	\end{tabular}
	\label{table: CNN}
\end{table}

\subsection{LeNet-5 Evaluation}
\begin{table}[!htbp]
	\centering
	\caption{LeNet-5 compression on Fashion MNIST dataset.}
	\begin{tabular}{c c c c c c c c }\hline
		&  		& CR	& Acc  		& Std 		& H 	& L  	 	& Storage	\\ 	\hline\hline
		& Original 	& 1x	& 89.86  	& 0.005	 	& 90.7  & 89.0	 	& 2.6 MB	\\ 	\hline
		& TRN~\cite{wang2018wide}  	& 13x  	& 88.10  	& 0.005 	& 89.0	& 87.1		& 0.2 MB	\\	\hline
		& NTRN  & 13x  	& 88.53 	& 0.004		& 89.4	& 87.8		& 0.2 MB	\\	\hline
		& TRN~\cite{wang2018wide}  	& 72x  	& 87.91  	& 0.006 	& 88.7	& 87.0		& 0.04 MB	\\	\hline
		& NTRN  & 72x  	& 88.47 	& 0.003		& 88.9	& 87.7		& 0.04 MB	\\	\hline
	\end{tabular}
	\label{table: LeNet5}
\end{table}

With promising results on monotypic layer compression, we conduct experiments using LeNet-5~\cite{lecun1998gradient} on the Fashion MNIST dataset to further evaluate the proposed NTRN, where  the LeNet-5 is made up of two convolutional and two fully-connected layers. The resized weight tensors from input to output layers are $\pmb{\mathcal{W}}_1 \in \mathbb{R}^{5\times 5 \times 1 \times 4 \times 5}$, $\pmb{\mathcal{W}}_2 \in \mathbb{R}^{5\times 5 \times 4 \times 5 \times 5 \times 10}$, $\pmb{\mathcal{W}}_3 \in \mathbb{R}^{5\times 10 \times 5 \times 5 \times 8 \times 8\times 8}$ and $\pmb{\mathcal{W}}_4 \in \mathbb{R}^{8\times 8 \times 8 \times 10}$. As shown in Table~\ref{table: LeNet5}, the proposed NTRN is superior to the TRN at both 13x and 72x CRs. Herein, the ranks are $\{3, 10, 30, 8\}$ for 13x CR, and $\{3, 8, 10, 5\}$ for 72x CR. 

\subsection{VGG-11 Evaluation}
Furthermore, the developed NTRN is examined by VGG-11 on two datasets, namely, Cifar-10 and Fashion MNIST. The VGG-11 consists of eight convolutional and three fully-connected layers. For the Cifar-10 dataset, the weight tensors from input to output layers are $\pmb{\mathcal{W}}_1 \in \mathbb{R}^{3\times 3\times 3\times 4 \times 4 \times 4 }$, $\pmb{\mathcal{W}}_2 \in \mathbb{R}^{3\times 3\times 4 \times 4 \times 4 \times 2 \times 4\times 4 \times4}$, $\pmb{\mathcal{W}}_3 \in \mathbb{R}^{3\times 3\times 2 \times 4\times 4 \times4 \times  4 \times 4 \times 4 \times 4}$, $\pmb{\mathcal{W}}_4 \in \mathbb{R}^{3\times 3\times 4 \times 4\times 4 \times4 \times  4 \times 4 \times 4 \times 4}$, $\pmb{\mathcal{W}}_5 \in \mathbb{R}^{3\times 3\times 4 \times 4\times 4 \times4 \times  4 \times 4 \times 4 \times 4 \times 2}$, $\pmb{\mathcal{W}}_6 =\pmb{\mathcal{W}}_7=\pmb{\mathcal{W}}_8 \in \mathbb{R}^{3\times 3\times 4 \times 4\times 4 \times4 \times  2\times 4 \times 4 \times 4 \times 4 \times 2}$, $\pmb{\mathcal{W}}_9 \in \mathbb{R}^{ 4 \times 4\times 4 \times4 \times  2 \times 4 \times 4 \times 4 \times 4\times 2}$, $\pmb{\mathcal{W}}_{10} \in \mathbb{R}^{ 4 \times 4\times 4 \times4 \times  2 \times 4 \times 4 \times 4 \times 4}$ and $\pmb{\mathcal{W}}_{11}\! \in\! \mathbb{R}^{ 4 \times 4\times 4 \times4 \times  10}$, respectively. Excluding the first weight tensor $\pmb{\mathcal{W}}_1 \in \mathbb{R}^{3\times 3\times 1\times 4 \times 4 \times 4 }$ for the Fashion MNIST dataset, the dimensions of the others are kept the same as those in Cifar-10. The ranks are set to $\{8,\! 25,\! 40,\! 50,\! 70,\! 70,\! 70, \!70,\! 15, \!15, \!5\}$ for 9x CR. By contrast, the ranks are $\{8,\! 25,\! 30,\! 40,\! 50,\! 50,\! 50,\! 50,\! 10,\! 10,\! 5\}$ at 17x CR.

\begin{table}[!htbp]
	\centering
	\caption{VGG-11 compression on Cifar-10 dataset.}
	\begin{tabular}{c c c c c c c c}\hline
		&  		& CR	& Acc  		& Std 		& H 	& L  		& Storage	\\ 	\hline\hline
		& Original 	& 1x	& 80.40  	& 0.005	 	& 81.0  & 79.4		& 36.6 MB	\\ 	\hline
		& TRN~\cite{wang2018wide}  	& 9x  	& 76.44   	& 0.006 	& 77.2	& 75.4		& 4.2 MB	\\	\hline
		& NTRN  & 9x  	& 77.87  	& 0.002		& 78.2	& 77.6		& 4.2 MB	\\	\hline
		& TRN~\cite{wang2018wide}  	& 17x  	& 75.67  	& 0.005 	& 76.6	& 75.1		& 2.2 MB	\\	\hline
		& NTRN  & 17x  	& 77.50	 	& 0.007		& 78.5	& 76.3		& 2.2 MB	\\	\hline
	\end{tabular}
	\label{table: VGG11_Cifar10}
\end{table}

Table~\ref{table: VGG11_Cifar10} lists the results on the Cifar-10 dataset. It is known that NTRN demonstrates overwhelming superiority compared with TRN. Specifically, the average accuracy by the proposed NTRN increases by 1.4\% and 1.8\% at 9x and 17x CRs, respectively. The corresponding results for the Fashion MNIST are tabulated in Table~\ref{table: VGG11_FMNIST}. Compared with TRN, the proposed NTRN effectively mitigates the loss of accuracy caused by compression.

\begin{table}[!htbp]
	\centering
	\caption{VGG-11 compression on Fashion MNIST dataset.}
	\begin{tabular}{c c c c c c c c }\hline
		&  		& CR	& Acc  		& Std 		& H 	& L  		& Storage	\\ 	\hline\hline
		& Original 	& 1x	& 92.93  	& 0.002	 	& 93.2  & 92.7		& 36.7 MB	\\ 	\hline
		& TRN~\cite{wang2018wide}  	& 9x  	& 91.81   	& 0.002 	& 92.0	& 91.6		& 4.2 MB	\\	\hline
		& NTRN  & 9x  	& 92.02  	& 0.002		& 92.3	& 91.7		& 4.2 MB	\\	\hline
		& TRN~\cite{wang2018wide}  	& 17x  	& 91.41  	& 0.003 	& 91.7	& 90.8		& 2.3 MB	\\	\hline
		& NTRN  & 17x  	& 91.79 	& 0.003		& 92.2	& 91.2		& 2.3 MB	\\	\hline
	\end{tabular}
	\label{table: VGG11_FMNIST}
\end{table}

\section{Conclusion}\label{sec: conclusion}
In this paper, we have proposed a novel network compression technique, termed as NTRN, where the weight arrays at fully-connected and convolutional layers are compressed by TR format. Different from the conventional TRN, a nonlinear activation function is added after tensor contraction and convolution operations inside the compressed layer. The proposed NTRN enables to enhance accuracy, as compared to the state-of-the-art TRN compression method. The superior performance of our NTRN has been verified using image classification task by different DNN's architectures, such as MLP, LeNet-5 and VGG-11 on MNIST, Fashion MNIST and Cifar-10 datasets. We believe that the proposed NTRN can be potentially used for embedded systems because of its effectiveness to achieve ultra-low memory cost.

\ifCLASSOPTIONcaptionsoff
  \newpage
\fi

\bibliographystyle{References/IEEEtran}
\bibliography{References/IEEEabrv,References/mybibfile}

\end{document}